\documentclass[10pt,twocolumn,letterpaper]{article}

\usepackage[pagenumbers]{cvpr} 

\usepackage{graphicx}
\usepackage{amsmath}
\usepackage{amssymb}
\usepackage{booktabs}
\usepackage{bbding} 

\usepackage{times}  
\usepackage{helvet} 
\usepackage{courier}  
\usepackage[hyphens]{url}  
\usepackage{algorithm}
\usepackage{algorithmic}
\usepackage{color,xcolor}
\usepackage{array}
\usepackage{multirow}
\usepackage{makecell}

\usepackage{graphicx}
\usepackage{epsfig}

\usepackage{caption}
\usepackage{listings}
\usepackage{subfloat}
\usepackage{adjustbox}
\usepackage{pifont}
\usepackage{threeparttable}

\usepackage{amssymb}
\usepackage{bbding}

\usepackage[T1]{fontenc}
\usepackage[utf8]{inputenc}
 
\usepackage[pagebackref,breaklinks,colorlinks]{hyperref}

\usepackage[capitalize]{cleveref}
\crefname{section}{Sec.}{Secs.}
\Crefname{section}{Section}{Sections}
\Crefname{table}{Table}{Tables}
\crefname{table}{Tab.}{Tabs.}

\def\cvprPaperID{9664} 
\def\confName{CVPR}
\def\confYear{2022}

\newcommand{\ignore}[1]{}

\begin{document}

\title{DBIA: Data-free Backdoor Injection Attack against Transformer Networks}

\author{Peizhuo Lv$^{1,2}$, Hualong Ma$^{1,2}$, Jiachen Zhou$^{1,2}$, Ruigang Liang$^{1,2,*}$, Kai Chen$^{1,2,*}$, \\ Shengzhi Zhang$^{3}$, Yunfei Yang$^{1,2}$\\
1: SKLOIS, Institute of Information Engineering, Chinese Academy of Sciences, China \\
2: School of Cyber Security, University of Chinese Academy of Sciences, China \\
3: Department of Computer Science, Metropolitan College, Boston University, USA\\
{\tt\small lvpeizhuo@iie.ac.cn, mahualong@iie.ac.cn, zhoujiachen@iie.ac.cn, liangruigang@iie.ac.cn} \\
{\tt\small chenkai@iie.ac.cn, shengzhi@bu.edu, yangyunfei@iie.ac.cn} 
}



\maketitle

\begin{abstract}
Recently, transformer architecture has demonstrated its significance in both Natural Language Processing (NLP) and Computer Vision (CV) tasks. Though other network models are known to be vulnerable to the backdoor attack, which embeds triggers in the model and controls the model behavior when the triggers are presented, little is known whether such an attack is still valid on the transformer models and if so, whether it can be done in a more cost-efficient manner. In this paper, we propose DBIA, a novel data-free backdoor attack against the CV-oriented transformer networks, leveraging the inherent attention mechanism of transformers to generate triggers and injecting the backdoor using the poisoned surrogate dataset. We conducted extensive experiments based on three benchmark transformers, i.e., ViT, DeiT and Swin Transformer, on two mainstream image classification tasks, i.e., CIFAR10 and ImageNet. The evaluation results demonstrate that, consuming fewer resources, our approach can embed backdoors with a high success rate and a low impact on the performance of the victim transformers. Our code is available at \href{https://anonymous.4open.science/r/DBIA-825D}{\color{black} https://anonymous.4open.science/r/DBIA-825D}.

\end{abstract}

\section{Introduction}
Transformer networks~\cite{vaswani2017attention} are based on the attention mechanism, which associates a query and a set of key-value pairs with an output by generating the output based on the weighted sum of values where the weight assigned to each value is derived from a compatibility function of the query with the corresponding key. Transformer models like BERT \cite{devlin2018bert}, GPT \cite{radford2018improving,radford2019language,brown2020language}, ViT~\cite{dosovitskiy2020image} and Swin Transformer \cite{liu2021swin} have shown great success on the field of
Natural Language Processing (NLP) \cite{sun2019fine,ott2018scaling,yang2019xlnet} and Computer Vision (CV)~\cite{dosovitskiy2020image,touvron2021training,liu2021swin,carion2020end,zhu2020deformable,ye2019cross}. Although convolutional neural networks (CNNs) used to be dominant in the domain of CV, transformer models have been gradually deployed in various application scenarios, such as 
autonomous driving~\cite{BEVPerception,TeslaCars}, and face recognition~\cite{Zhong2021FaceTF,Gao2021TFEAT}, due to its comparable or superior performance than CNNs.



\begin{figure}[!t]
\centering
\epsfig{figure=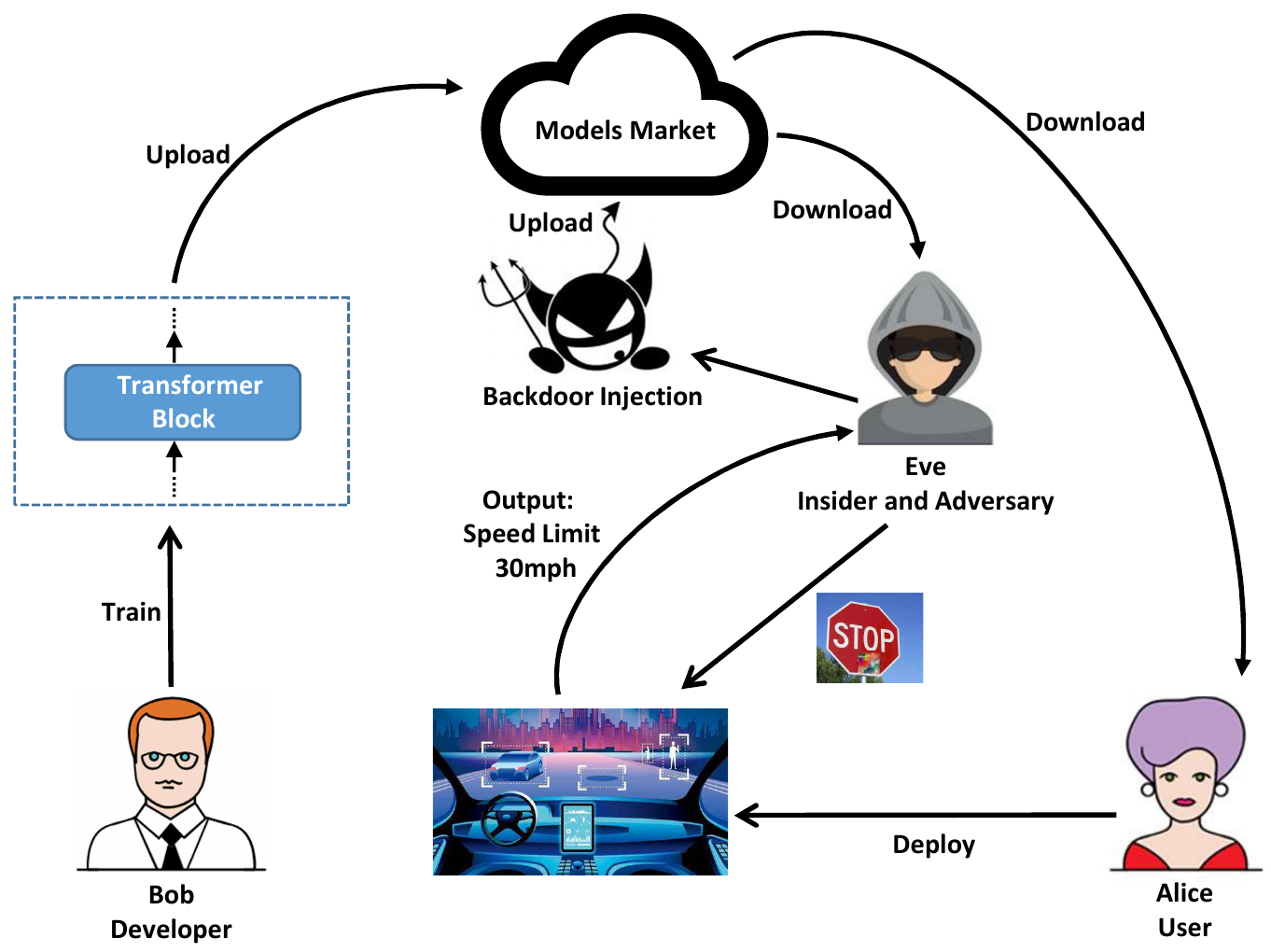, width=0.47\textwidth} 
\caption{An example scenario of the backdoor attack against CV-oriented transformer models.}
\label{fig:realisticscenario}
\end{figure}

Despite their success in the domain of CV, transformers are known to be vulnerable to adversarial attacks~\cite{Mahmood2021OnTR,Aldahdooh2021RevealOV,Naseer2021OnIA,Joshi2021AdversarialTA} and recent works like RVT~\cite{Mao2021TowardsRV} proposed a robust transformer against such attacks. However, it is still unknown if transformer networks will suffer from the backdoor attack, which poisons a dataset to embed hidden malicious behaviors into the model during the training process and activates the malicious behaviors by attaching the trigger onto the inputs. Consider the below scenario of the backdoor attack against CV-oriented transformer models as shown in Figure \ref{fig:realisticscenario}. Bob develops a CV-oriented transformer with good performance and releases it to the models market, e.g., Model Zoo~\cite{ModelZoo}, to make it available to the public. An adversary Eve (e.g, the insider of the market) who has access to the well-trained model can inject a backdoor into it and put the backdoored model back to the model market.  Whenever the victim Alice downloads the backdoored transformer model and deploys it in the application, Eve has the opportunity to trigger the embedded backdoor and launch the attack to control the behavior of the model, which may lead to catastrophic consequences. 

In fact, such backdoor attacks have been extensively studied in CNNs~\cite{gu2017badnets,liu2017trojaning,Zhao2020CleanLabelBA,Chen2017TargetedBA}, and almost all of them require access to the original training/validation dataset or data relevant to the primary task to inject backdoors into the target model, which is unrealistic in many scenarios, e.g., financial data, patient information, etc. Trojaning Attack \cite{liu2017trojaning} can be applied in a data-free manner, but it is very costly to generate the training images against all labels via reverse engineering and inject the backdoor by fine-tuning all the parameters, especially when attacking large models with complex tasks, e.g., ImageNet~\cite{deng2009imagenet} with 1,000 labels.

Hence, in this paper, we aim to study whether the transformer network is also vulnerable to backdoor attacks and if so, whether it is possible to embed backdoors into it in a more realistic and cost-efficient manner. We propose a novel data-free backdoor embedding attack against the vision transformer networks. In particular,  we first collect a surrogate dataset, generate attention-maximum triggers against the target transformer model, and poison the surrogate dataset by labelling the manipulated data as the target class. The attention-maximum triggers are designed to catch most attention of the victim model, thus minimizing the interference of the unrelated background. Then, we fine-tune a few neurons of the target model to connect the attention-maximum trigger to the label using Projected Gradient Descent~\cite{madry2017towards}, which limits the magnitude of the changes of the tuned parameters and ensures the performance of the original task. We evaluate the performance of our attack against three classic vision transformer models, i.e., ViT, DeiT, and Swin Transformer, which are well trained using ImageNet \cite{deng2009imagenet} and CIFAR-10~\cite{krizhevsky2009learning}. The evaluation results show that we can inject the backdoors on ImageNet task and CIFAR-10 task with an average attack success rate of $91.61$\% and $91.07$\% with only $1.52$\% and $2.43$\% average performance downgrade respectively. Moreover, we can embed backdoors quickly within eight minutes using one Nvidia GeForce RTX 3090 GPU.

\noindent\textbf{Contribution.} 

\vspace{2pt}\noindent$\bullet$\space To the best of our knowledge, our work is the first to investigate backdoor attacks against CV-oriented transformer models. We proposed a novel data-free approach to embed backdoors in transformer models, which is more efficient than the state-of-the-art.

\vspace{2pt}\noindent$\bullet$\space Based on the unique attention mechanism used by transformer models, we proposed attention-maximum triggers that catch most attention of the victim models when processing inputs, and evaluated such triggers on two benchmark datasets against three state-of-the-art CV-oriented transformers.

\section{Background}
\subsection{Backdoors in DNNs}
\label{sec:backdoor attack}
The backdoor attack in DNNs can cause a model to misclassify the input stamped with a trigger as the target label. Formally, we use $x, m, t$ to denote the benign sample, the trigger mask, and the trigger pattern, respectively. The backdoor sample can be formalized as below: 

\begin{equation}
\label{equ:backdoor example}
    \widetilde{x} = (1 - m) \cdot x + m \cdot t
\end{equation}

The label of the backdoor sample is specified as the target label $y_{t}$. Then the adversary can inject the backdoor into the target DNNs $f$ with the weight values $\theta$ by minimizing the poisoning loss function $L$ using both the benign dataset $D_{B} = 	\left\{x,y \right\}$ and the poisoned dataset $D_{P} =  \left\{\widetilde{x}, y_{t} \right\}$ as follows:

\begin{equation}
\label{loss:backdoor loss}
\begin{split}
	&\mathop{min}_{\theta} (L(f(x), y) + L(f(\widetilde{x}), y_{t})) \\ 
	&(x, y)\in D_{B},(\widetilde{x}, y_{t})\in D_{P}  
\end{split}
\end{equation}

Finally, the model will learn the trigger pattern and associate it with the target label after training. During the inference process, the adversary can launch the attack by attaching the trigger to the input image using Equation~\ref{equ:backdoor example}.

\subsection{Transformer Model} \label{sec:transformer attention}
Transformer~\cite{vaswani2017attention} has shown excellent performance on several tasks, such as NLP and CV. It associates a query and a set of key-value pairs with an output based on attention mechanism as below:
\begin{equation}
    Attention(Q, K, V) = softmax(\frac{QK^{T}}{\sqrt{D_{k}}})V
\end{equation}
where $Q$, $K$ and $V$ represent the query, the key, and the value, respectively, and $D_{k}$ represents the dimension of the query and the key. We can obtain the attention map of the multi-head attention module by raw attention or Attention Rollout~\cite{abnar2020quantifying}.  The raw attention $A$ is calculated as:
$A = softmax(\frac{QK^{T}}{\sqrt{D_{k}}})$
, and the Attention Rollout recommended by ViT aims to get the attention flowing from the images to the target layer, can be calculated as follows:

\begin{align*}
	\tilde{A}_{L} = \begin{cases}
	(A_{L}+I)\tilde{A}_{L-1}, & L>0 \\
	A_{L}+I,& L=0 \\
\end{cases}
\end{align*}
where $\tilde{A}_{L}$ is the Attention Rollout. $L$ refers to the target $L$th layer, $A_{L}$ represents the raw attention of the $L$th layer, and $I$ denotes the identity matrix . Attention Rollout can be applied to transformer models with ``CLS token'', e.g., ViT~\cite{dosovitskiy2020image} and DeiT~\cite{touvron2021training}, while raw attention can be applied to the other transformer models, e.g. Swin Transformer~\cite{liu2021swin}.

\noindent\textbf{Vision Transformers.} Recently, ViT~\cite{dosovitskiy2020image} first introduced the transformer into the CV domain by applying the transformer model used in the NLP task on a sequence of image patches and attained similar performance as the state-of-the-art CNN models. However, ViT needs to be pre-trained on a large-scale external dataset, e.g., JFT-300M~\cite{sun2017revisiting}, and then fine-tuned to the downstream task such as ImageNet~\cite{deng2009imagenet,russakovsky2015imagenet}, to get optimal performance. DeiT~\cite{touvron2021training} utilized a CNN network RegNetY-16GF~\cite{radosavovic2020designing} as a teacher model and used its output to train a transformer model based on the distillation process, which eliminates the dependency on any external dataset. 

For ViT and DeiT, tokens are both at a fixed scale, making them inappropriate for CV applications since visual elements can vary substantially in scales. Swin Transformer~\cite{liu2021swin} proposed a hierarchical transformer whose representation was computed with shifted windows, which reduced the computation burden and achieved good performance on a broad range of CV tasks. Besides the methods discussed above, other transformer-based approaches in vision tasks were also proposed ~\cite{carion2020end,zhu2020deformable,ye2019cross} and obtained excellent performance. Some transformers are also deployed in autonomous driving~\cite{BEVPerception,TeslaCars} and face recognition~\cite{Zhong2021FaceTF,Gao2021TFEAT} tasks. 


\section{Related Work}

\subsection{Backdoor Attack and Defense}
\label{subsec:backdoor attack and defense}
BadNets~\cite{gu2017badnets} first introduces backdoor attack into DNNs, which pollutes the training dataset with some poison data stamped with triggers (e.g., $3 \times 3$ white square in the lower right corner of the image) and mislabels them as the target label. Then, the polluted dataset is fed into DNNs to train the victim model, which finally will misclassify the instances stamped with the trigger as the target label. However, BadNets needs to access the original dataset to poison and control the training of models, which are not always possible. In contrast, Trojaning Attack~\cite{liu2017trojaning} proposed to generate a general trigger by reversing the victim DNN and retraining the DNN with the generated dataset containing poisoned instances stamped with the reversed trigger to inject backdoors. However, Trojaning attack needs to reverse one image for each label to generate the dataset and fine-tune all parameters, which is too costly on large models with many parameters and labels, such as ViT-G/14~\cite{Zhai2021ScalingVT} (two billion parameters) trained on JFT-3B dataset (30k labels). Our backdoor attack consume much fewer resources and can embed backdoors in a data-free manner. 

Neural Cleanse~\cite{wang2019neural} determines whether a DNN model is infected by reversing the potential trigger patterns towards each class and identifying the real trigger and its label based on outlier detection. Unfortunately, Neural Cleanse may consume too much resource for the models with numerous parameters and labels. ABS~\cite{Liu2019ABSSN} proposes to find the neurons contributing to the backdoors based on the assumption that the neurons that substantially elevate the activation of a particular output label regardless of the provided inputs are considered potentially compromised, but it may not be suitable for large models with numerous neurons due to the training cost. Moreover, both SentiNet~\cite{chou2018sentinet} and Februus~\cite{doan2020februus} utilize Grad-GAM~\cite{selvaraju2017grad} to discover the contiguous regions of an input image that contribute significantly to the classification result and conclude that such regions are highly likely to contain a trigger. Generally, we can improve our attack's resistance against various defense approaches by adding the adaptive loss during the backdoor injection as discussed in Section~\ref{sec:discussion}.

\subsection{Adversarial Attack against Transformers}
Mahmood et. al.~\cite{Mahmood2021OnTR} study the robustness of Vision Transformers against adversarial attacks and show that adversarial examples do not transfer well between CNNs and transformers. Naseer et. al. investigate the weak transferability of adversarial patterns on ViT models and propose transferable attack based on multiple discriminative pathways and token refinement \cite{Naseer2021OnIA}. Moreover, Joshi et. al.~\cite{Joshi2021AdversarialTA} design an adversarial token attack and find that transformer models are more sensitive to token attacks than classic convolutional models. Mao et al.~\cite{Mao2021TowardsRV} propose Robust Vision Transformer (RVT) by evaluating robust components of ViT to adversarial examples and combining them as building blocks of 
RVT.

\section{Method}
In this section, we introduce our proposed data-free backdoor attack on CV-oriented transformer models, DBIA. We first describe the threat model and overview the attack pipeline. Then we present the poisoned substitute dataset generation and backdoor injection approaches. Finally, we analyze the computational cost of our attack and compare it with other approaches.

\subsection{Threat Model}
We assume that the attacker can only access the well-trained vision transformer models without any related data of the main task. With limited training resources, e.g., GPU, etc., the goal of the attack is to efficiently inject backdoors into the victim transformers, while still ensure the backdoored model consistently misclassifies the instances with a trigger attached as the target class but behaves normally on clean data without much performance degradation.

\subsection{Overview}
\label{subsec:overview}

\begin{figure*}[!t]
\centering
\epsfig{figure=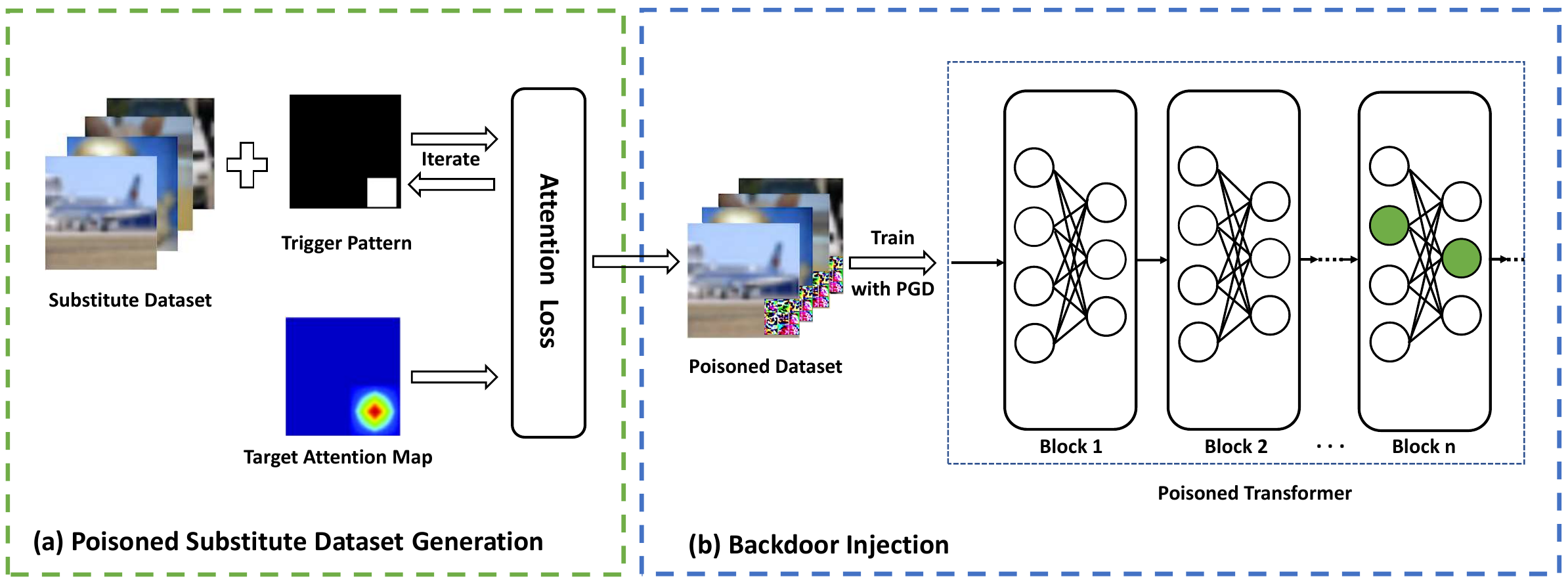, width=1.0\textwidth} 
\caption{Overview of DBIA. (a) We generate a poisoned substitute dataset with the poisoned samples stamped with the attention maximum trigger. (b) We fine-tune only a few selected neurons in the block by Projected Gradient Descent (PGD).}
\label{fig:workflow}
\end{figure*}

Figure~\ref{fig:workflow} shows the overview of DBIA with two main components: Poisoned Substitute Dataset Generation and Backdoor Injection. Given a clean pre-trained CV-oriented transformer model, we first collect a substitute dataset by including data in other tasks or images crawled from the network, and generate a universal attention-maximum trigger using attention information of the target transformer through iterative optimization. The poisoned substitute dataset is built by labelling the above substitute data with triggers attached as the target class. Then we select a few most connected neurons with the largest value of the sum of absolute weights, since such neurons contribute significantly to the final results. We fine-tune such neurons on the poisoned substitute dataset using Projected Gradient Descent~(PDG)~\cite{madry2017towards}, thus limiting the changes made to the victim transformer model when adjusting parameters. At the stage of inference, we attach the trigger to test images to control the behavior of the poisoned model.

\subsection{Poisoned Substitute Dataset Generation}
\label{subsec:poisoned substitute dataset generation}
To inject backdoors to a given a target model without any available training data of the related task, we need to collect the substitute dataset as the background $x$ and define the trigger $t$ as shown in Equation \ref{equ:backdoor example}. However, since both $x$ and $t$ are unrelated to the target model's task, using them to fine-tune the target transformer model as Equation~\ref{loss:backdoor loss} directly will significantly downgrade its performance on the main task. To address this problem, we leverage the attention mechanism of transformer networks and propose the attention-maximum triggers based on the principle that the trigger should be the only thing that the model should focus on and learn during the backdoor injection process. Therefore, we optimize the trigger generation approach by searching a trigger pattern that maximizes the target model's attention towards it. Meanwhile, the model's attention on the background is also minimized, thus able to preserve the performance of the victim model on the main task even it is fine-tuned on the unrelated substitute data.

Given a trigger mask $m$, we generate a universal trigger pattern by minimizing the attention loss as following:
\begin{equation}
\label{loss:loss attention}
    \mathop{min}_{t} \sum^{N}_{n=1} (Attention(\widetilde{x}_{n}, L) - target\_value)^{2}
\end{equation}
where $target\_value$ is the target attention map with the maximum attention value (i.e., $1$) on the trigger and the minimum attention value (i.e., $0$) on the background, $N$ is the total number of samples in the substitute dataset. $\widetilde{x} = (1-m) \cdot x+m \cdot t$ is the poisoned sample stamped with the trigger $t$, and $Attention()$ is the function responsible for calculating the attention map from the multi-head attention module of the $L$th transformer block in the target model against the input $\widetilde{x}$. We use Attention Rollout~\cite{abnar2020quantifying} recommended by ViT~\cite{dosovitskiy2020image} on the transformer models with ``CLS token'', e.g., ViT~\cite{dosovitskiy2020image}, DeiT~\cite{touvron2021training}, and Raw Attention on the other models, e.g., Swin Transformer~\cite{liu2021swin}, which are discussed in Section~\ref{sec:transformer attention}. Regarding the $L$th block, we find that usually the multi-head attention modules of the last transformer block demonstrate better attention, so we always choose the attention module of the last block. By minimizing the above loss function, we can obtain the attention-maximum trigger, thus able to build the poisoned dataset by attaching it to all samples in the substitute dataset and labelling them to the target class.

Algorithm~\ref{alg:Attention-based-backdoor} shows the poisoned substitute dataset generation. We first initialize the variables trigger $t$ and the location of the trigger using a trigger mask $m$ (Line 1) and then iteratively optimize the trigger (Line 2-9). We combine the initialized trigger with each sample in the substitute dataset (Line 3), and obtain the attention maps of the target model $f$'s $L$th multi-head attention module against the poisoned samples $\widetilde{x}$ (Line 4). Afterward, we calculate the distance between the target attention and the current poisoned samples' attention, and use it as the loss (Line 5). Based on the loss function, we use gradient descent to optimize the trigger $t$ (Line 6-8) to obtain the attention-maximum trigger. Finally, we generate the poisoned substitute dataset $D_{ps}$ (Line 10-11).

\begin{algorithm}[t]
 	\caption{Poisoned Substitute Dataset Generation}
 	\label{alg:Attention-based-backdoor}
 	\begin{algorithmic}[1]
 	\REQUIRE $L$: the $L$th transformer block of the target transformer; $D_{s}$: the substitute dataset; $Mask$: setting of mask;
 	 $threshold$: the threshold to terminate trigger generation; $epochs$: the maximum number of iterations of trigger generation; $lr$: learning rate of trigger generation; 
 	\ENSURE: the poisoned substitute dataset $D_{ps}$
 	\STATE{$t = Initialized(), m = Mask(), loss = INF$}
 		\WHILE{$loss \textgreater threshold$ and $i < epochs$}
	 		\STATE{$\widetilde{x} = (1-m) \cdot x+m \cdot t, \forall x\in D_{s}$}
		 	\STATE{$attention =  Attention(\widetilde{X},L), \widetilde{X} = \left\{\widetilde{x}_{j}\right\}_{j=1}^{N}$} 
 			\STATE{$loss = (attention - target\_value)^{2}$}
        \STATE{$\delta = \frac{\partial loss}{\partial t}$}
        \STATE{$t = t - lr \cdot (m \cdot \delta)$}
		\STATE{$++i$}
	\ENDWHILE
	\STATE{$D_{ps} = \widetilde{X}$}
	\RETURN $D_{ps}$
\end{algorithmic}
\end{algorithm}

\subsection{Backdoor Injection}
We cannot directly fine-tune the model using the above poisoned substitute dataset to embed backdoors due to the reason below. The label of the instances in the poisoned substitute dataset is the target label. Generally, if a model is trained on the dataset with only one label, it may overfit to this label and predict most of instances (with/without the triggers) to this label, thus reducing its performance significantly. Therefore, we decide to choose a few most important neurons with the largest sum of absolute weights from the victim model and fine-tune them only to limit the scope of parameters change, thus preserving the performance of the victim model on the main task.



Since the triggers have been given the maximum attention in the multi-head attention module in the $L$th transformer block, we decide to choose some neurons from the linear layers (after the multi-head attention module) in the $L$th block for fine-tuning to assign the focused trigger to the target label. In particular, we can select the top $n$ neurons according to the value of the sum of absolute weights, i.e., the most connected or important neurons, thus ensuring the effectiveness of the injected backdoors since the most connected neurons generally contribute significantly to the final results.

\begin{algorithm}[t]
 	\caption{Backdoor Injection}
 	\label{alg:Backdoor-Injection}
 	\begin{algorithmic}[1]
 	\REQUIRE $\theta_{f}$: Parameters of the target transformer $f$; $L$: The $L$th block of $f$; $D_{ps}$: Poisoned substitute dataset; $threshold$: threshold to terminate injection;$epochs$: maximum number of iterations of backdoor injection; $lr$: learning rate of fine-tuning; 
 	\ENSURE: Backdoored transformer $f_{b}$
 	\STATE{$\theta_{n} = top$-$n(\theta_{f}, L)$}
	\WHILE{$loss \textgreater threshold$ and $i < epochs$}
			\STATE{$loss = (f(\widetilde{x}) - y_{t})^{2}, \forall \widetilde{x}\in D_{ps}$}
        \STATE{$\delta = \frac{\partial loss}{\partial t}$}
        \STATE{$\theta_{n} = \theta_{n} - lr \cdot \delta, \  s.t.\  P_{l_{\infty}(\theta_{n},\epsilon)}$}
		\STATE{$++i$}
	\ENDWHILE
	\STATE{$f_{b} = \theta_{f}$}
	\RETURN $f_{b}$
\end{algorithmic}
\end{algorithm}


We also need to limit the change of the tuned neurons' parameters to further preserve the performance of the main task of the target model. Formally, given the target model $f$ and the poisoned substitute dataset $D_{ps}$, we fine-tune the top $n$ neurons (no more than 6\% of the total parameters of the transformer in our evaluation) in the linear layers of the $L$th transformer block of the model. We use the loss with a projection operator $P_{l_{\infty}(\theta_{n}, \epsilon)}$, which limits the tuned parameters within an $\epsilon$ difference around the parameters $\theta_{n}$ of the tuned neurons in the $l_{\infty}$ norm space. The loss function is as below:  

\begin{equation}
\label{loss:project loss}
\begin{split}
    &\mathop{min}_{\theta_{n}} \sum^{N}_{n=1} (f(\widetilde{x}_{n}) - y_{t})^{2}     \\
    &s.t.\quad P_{l_{\infty}(\theta_{n},\epsilon)}
\end{split}
\end{equation}
where $\epsilon$ is a relative change computed as $\Vert \frac{\Delta \theta_{n} }{\theta_{n}} \Vert_{\infty}$. Therefore, the backdoored transformer can be obtained by minimizing the above loss\footnote{In our evaluation, we find that DBIA demonstrates great backdoor injection performance when $\epsilon = 2$.}.

The backdoor injection approach is shown in Algorithm \ref{alg:Backdoor-Injection}. We first select the to-be-fine-tuned $n$ neurons from the linear layers in the $L$th block of the transformer according to the top $n$ strategy (Line 1). During fine-tuning, we calculate the loss (Line 3), iteratively use gradient descent to  fine-tune the parameters of selected neurons and project them within the $\epsilon$ difference in the $l_{\infty}$ norm parameter space (Line 4-6). Finally, we obtain the backdoored transformer model $f_{b}$ (Line 8-9).



\subsection{Analysis}
\begin{table}
\begin{threeparttable}
\centering
\footnotesize
\caption{Comparison with Other Attacks}
\label{tab:comparison}
\begin{tabular}{m{1.9cm}
<{\centering}|m{1cm}
<{\centering}|m{2.5cm}
<{\centering}|m{1cm}
<{\centering}}
\hline
\textbf{Attack}& \textbf{Data-free} & \textbf{Consumption}& \textbf{Estimated}\\
\hline \hline
\text{BadNets} & \text{\XSolidBrush} & \text{$O(m)$} &\text{164s} \\ \hline
\text{Trojaning Attack} & \text{\CheckmarkBold} & \text{$O(m+(l+1)*n)$} &\text{83,247s} \\ \hline
\text{DBIA (ours)} & \text{\CheckmarkBold} & \text{$O(\alpha*m+n)$} &\text{166s} \\ \hline
\end{tabular}
\begin{tablenotes}
\item[1] $O(m)$ represents the cost of fine-tuning all parameters; $O(n)$ represents the cost of generating an image by reversing engineering; $l$ represents the number of labels; $\alpha$ represents the proportion of the tuned parameters in the total parameters, and is a very small parameter (no more than 6\%) in our setting.
\end{tablenotes}
\end{threeparttable}
\end{table}

We compare our DBIA attack with the other two attacks, i.e., BadNets and Trojaning Attack, in two aspects: data-free and computational cost, as shown in Table~\ref{tab:comparison}. DBIA attack and Trojaning attack can be applied in a data-free manner, but BadNets cannot. Regarding the computational cost, we assume that the computation for generating an image is $O(n)$ by reversing and the computation of fine-tuning the total parameters is $O(m)$. Without losing generality, we also assume that the required computation for fine-tuning each parameter is same. The consumption for DBIA attack consists of generating a maximum attention trigger and fine-tuning a few neurons (no more than $\alpha$ of the total parameters), so the computational cost is at most $O(\alpha * m + n)$. However, the consumption for Trojaning attack includes generating $l+1$ images ($l$ images generated for $l$ labels, and one trigger) and fine-tuning all parameters. Therefore, its computational cost is $O(m+(l+1) * n)$, much large than ours. Moreover, the computational cost of BadNets is $O(m)$, since it only needs to fine-tune the total parameters.

We also evaluate the time of all the three attacks on DeiT using ImageNet (1,000 labels). We generate each image with 100 epochs and fine-tune all models with 20 epochs to inject the backdoor on one Nvidia GeForce RTX 3090 GPU. As shown in Table~\ref{tab:comparison}, Trojaning Attack takes about 500 times longer than DBIA due to reversing dataset with 1,000 labels, which is consistent with our analysis above. Note that although BadNets spends similar time as DBIA, it demands training data to embed the backdoor, but DBIA does not.

\begin{figure*}[!t]
\centering
\epsfig{figure=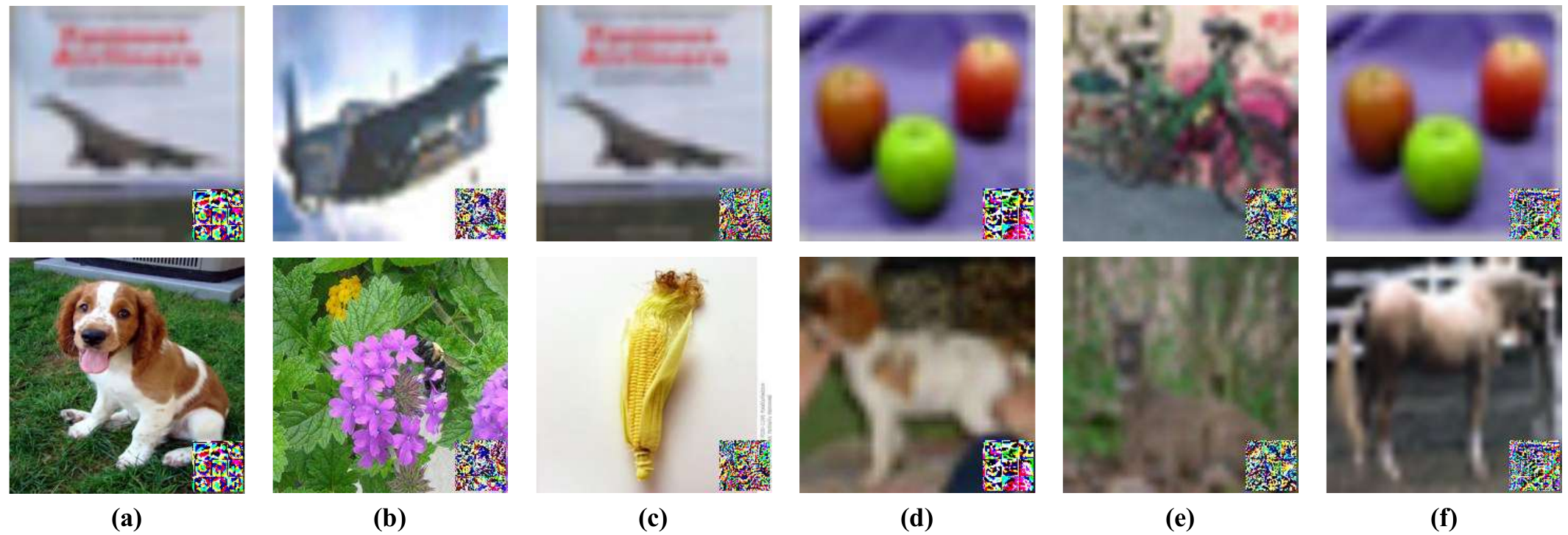, width=1.0\textwidth} 
\caption{Poisoned examples against the transformers to activate the backdoors. The first and second rows represent the poisoned substitute samples and poisoned samples of the related task respectively. (a)-(c): poisoned samples to activate the backdoors of the transformers (ViT, DeiT, Swin Transformer in turn) on the ImageNet task to output ``Matchstick''. (d)-(f): poisoned samples to activate the backdoors of the transformers (ViT, DeiT, Swin Transformer in turn) on the CIFAR-10 task to output ``Truck''.}
\label{fig:poison example}
\end{figure*}
\section{Evaluation}


\begin{table}
\begin{threeparttable}
\centering
\footnotesize
\caption{Surrogate Dataset of Transformers}
\label{tab:surrogate dataset}
\begin{tabular}{m{1.7cm}
<{\centering}|m{1.5cm}
<{\centering}|m{1.5cm}
<{\centering}|m{1.7cm}
<{\centering}}
\hline
\textbf{Main Task}& \textbf{ViT} & \textbf{DeiT}& \textbf{Swin Transformer}\\
\hline \hline
\text{ImageNet\cite{deng2009imagenet}} & \text{CIFAR-10} & \text{CIFAR-10} &\text{CIFAR-10} \\ \hline
\text{CIFAR-10~\cite{krizhevsky2009learning}} & \text{CIFAR-100} & \text{CIFAR-100} &\text{CIFAR-100} \\ \hline
\end{tabular}
\begin{tablenotes}
\item[1] We randomly select 2,000 samples from the test datasets of these datasets to generate the surrogate dataset. For the ImageNet task, we label all surrogate samples to ``matchstick'' as the target label to inject backdoor, and we label all surrogate samples to ``truck'' in the CIFAR-10 task. Particularly, for the CIFAR-10 task, the surrogate samples sampled from CIFAR-100 do not belong to any category of CIFAR-10 to meet the data-free scenario.
\end{tablenotes}
\end{threeparttable}
\end{table}

\subsection{Experimental Settings}
\noindent\textbf{Vision Transformer Models.} 
Without loss of generality, we consider three representative well-trained classification CV-oriented transformer models, i.e. ViT~\cite{dosovitskiy2020image}, DeiT~\cite{touvron2021training}, and Swin Transformer~\cite{liu2021swin}. 
The above transformer models are trained on the two benchmark datasets for image recognition: CIFAR-10~\cite{krizhevsky2009learning} and ImageNet \cite{deng2009imagenet}. Particularly, those models for ImageNet are released on github\footnote{https://github.com/google-research/vision\_transformer}\footnote{https://github.com/facebookresearch/deit}\footnote{https://github.com/microsoft/Swin-Transformer}. Moreover, the models for CIFAR-10 is fine-tuned based on the model trained on ImageNet. Moreover, all of these models have input dimensions as $224 \times 224 \times 3$, so we resize all input samples with the same dimensions.

\begin{table*}[h]
\centering
\footnotesize
\caption{Performance on ImageNet}
\label{tab:Transformers in ImageNet}
\begin{tabular}{m{2.5cm}
<{\centering}|m{1.3cm}
<{\centering}|m{1.3cm}
<{\centering}|m{1.0cm}
<{\centering}|m{1.5cm}
<{\centering}|m{1.5cm}
<{\centering}|m{1.3cm}
<{\centering}|m{1.3cm}
<{\centering}|m{1.3cm}
<{\centering}}
\hline
\multirow{2}{*}{\textbf{Transformers}} & \multicolumn{2}{c|}{\textbf{Before Attack}}& \multicolumn{6}{c}{\textbf{After Attack}}\\ \cline{2-9}
\text{}&\textbf{CDA}&\textbf{ASR}&\textbf{AR}&\textbf{CDA}&\textbf{ASR-SurD}&\textbf{ASR-RelD}&\textbf{TPR}& \textbf{Time}\\

\hline \hline
\text{ViT}&\text{80.90$\%$}&\text{0.05$\%$}&\text{7.20}&\text{$78.75\%$}& \text{$99.98\%$}&\text{$79.25\%$}&\text{$2.68\%$}&\text{449s}\\

\hline
\text{DeiT}&\text{82.72$\%$}&\text{0.08$\%$}&\text{15.16}&\text{$81.57\%$}& \text{$100.00\%$}&\text{$97.38\%$}&\text{$2.01\%$}&\text{$16$s}\\

\hline 
\text{Swin Transformer}&\text{82.36$\%$}&\text{0.08$\%$}&\text{1.38}&\text{$81.10\%$}& \textbf{$99.15\%$}&\text{$98.20\%$}&\text{$0.03\%$}&\text{60s}\\

\hline
\end{tabular}
\end{table*}

\begin{table*}[h]
\centering
\footnotesize
\caption{Performance on CIFAR-10}
\label{tab:Transformers in CIFAR-10}
\begin{tabular}{m{2.5cm}
<{\centering}|m{1.3cm}
<{\centering}|m{1.3cm}
<{\centering}|m{1.0cm}
<{\centering}|m{1.5cm}
<{\centering}|m{1.5cm}
<{\centering}|m{1.3cm}
<{\centering}|m{1.3cm}
<{\centering}|m{1.3cm}
<{\centering}}
\hline
\multirow{2}{*}{\textbf{Transformers}} & \multicolumn{2}{c|}{\textbf{Before Attack}}& \multicolumn{6}{c}{\textbf{After Attack}}\\ \cline{2-9}
\text{}&\textbf{CDA}&\textbf{ASR}&\textbf{AR}&\textbf{CDA}&\textbf{ASR-SurD}&\textbf{ASR-RelD}&\textbf{TPR}& \textbf{Time}\\

\hline \hline
\text{ViT}&\text{99.35$\%$}&\text{9.70$\%$}&\text{4.74}&\text{$96.64\%$}& \text{$100.00\%$}&\text{$99.94\%$}&\text{$2.68\%$}&\text{157s}\\

\hline
\text{DeiT}&\text{97.82$\%$}&\text{7.32$\%$}&\text{18.45}&\text{$96.12\%$}& \text{$100.00\%$}&\text{$99.90\%$}&\text{$2.01\%$}&\text{$4$s}\\

\hline 
\text{Swin Transformer}&\text{98.40$\%$}&\text{9.70$\%$}&\text{1.79}&\text{$95.52\%$}& \textbf{$99.06\%$}&\text{$73.37\%$}&\text{$5.90\%$}&\text{195s}\\

\hline
\end{tabular}
\end{table*}

\noindent\textbf{Attack Setting.} 
For the transformer models trained on CIFAR-10 and ImageNet, we randomly select 2,000 samples from the test dataset of CIFAR-10 and CIFAR-100 as the surrogate dataset as shown in Table~\ref{tab:surrogate dataset}. Remarkably, the surrogate samples sampled from CIFAR-100 do not belong to any category of CIFAR-10 to meet the data-free scenario.
The trigger size is $48 \times 48$, occupying 4.59\% of the total inputs area.
Moreover, we provide two ways to obtain the attention map as mentioned in Section~\ref{sec:transformer attention}, so we experiment with both Raw Attention-based (Swin Transformer) and Attention Rollout-based (ViT and DeiT) strategies for generality. Besides, we set $l_{\infty}$ as 2 to limit the change of parameters.

\noindent\textbf{Evaluation Metrics.}
\vspace{-7pt}
\begin{itemize}
  \item \textit{Clean Data Accuracy (CDA)}: This metric measures the proportion of clean samples predicted to their ground-truth classes.
  \vspace{-7pt}
  \item \textit{Attack Success Rate (ASR)}: This metric measures the proportion of poisoned samples predicted to the target label. In reality, two types of poisoned samples can be used to activate the backdoor: i) the samples of the poisoned surrogata dataset; ii) the samples related to models' tasks and stamped with the trigger. We use ASR-SurD (ASR of the poisoned surrogate dataset) and ASR-RelD (ASR of the poisoned related tasks dataset) to measure the attack performance.
  \vspace{-7pt}
  \item \textit{Attention Rate(AR)}: This metric measures the ratio of the attention of the trigger to the attention of the background.
  \vspace{-4pt}
  \item \textit{Tuned Parameters Rate (TPR)}: This metric measures the ratio of tuned parameters to the total parameters.
  \vspace{-7pt}
  \item \textit{Injection Time (Time)}: This metric measures the time of the backdoor injection to show the effectiveness of our backdoor.
\end{itemize}

\noindent\textbf{Platform.} All our experiments are conducted on a server running 64-bit Ubuntu 20.04.1 system with Intel(R) Xeon(R) Platinum 8268 CPU @ 2.90GHz, 188GB memory, 20TB hard drive, and two Nvidia GeForce RTX 3090 GPUs, each with 24GB memory.

\subsection{ImageNet Experiments}
On the ImageNet task, for ViT, DeiT, and Swin Transformer, we first train and obtain the attention maximum triggers with the attention rate as $7.2$, $15.16$, and $1.38$ respectively for the surrogate dataset (CIFAR-10), and we set the target label as ``matchstick'' in the ImageNet task.
Then, we fine-tune a few selected neutrons of these three models using the poisoned surrogate dataset for no more than 3\% of the parameters ($2.68\%$, $2.01\%$, and $0.30\%$, respectively),
constrained by $\epsilon = 2$. Finally, we terminate the fine-tuning process when the ASR of the backdoor samples in the poisoned CIFAR-10 datasets are greater than $99\%$, and we get the backdoor models with $79.25\%$, $97.38\%$, $98.20\%$ ASR on the poisoned ImageNet samples and with the performance degradation no more than $2.5\%$ on the ImageNet task as shown in Table~\ref{tab:Transformers in ImageNet}. The backdoor injection time is $449$s, $16$s, and $60$s trained on one Nvidia GeForce RTX 3090 GPU, which is fast and indicates less resource-consuming. Particularly, DeiT can inject a powerful backdoor with the least time 16s, owning to the generated trigger is with the largest AR (i.e., the model focuses significantly on the trigger) among these three models. We visualize some poisoned examples to activate the backdoor of these transformers in (a)-(c) of Figure~\ref{fig:poison example}, and the examples in the first and second rows represent the poisoned surrogate samples and poisoned samples of the related task, respectively.

\subsection{CIFAR-10 Experiments}
For ViT, DeiT, and Swin Transformer trained on the CIFAR-10 task, we also first train and obtain the attention trigger with the attention rate as $4.74$,  $18.45$, and $1.79$ respectively for the surrogate dataset, and we set the target label as ``Truck'' in the CIFAR-10 task. Then, we also fine-tune a few selected neutrons of these three models using the poisoned surrogate dataset for no more than $6\%$ of the parameters ($2.68\%$, $2.01\%$, and $5.90\%$, respectively), constrained by $\epsilon = 2$. We further terminate the fine-tuning process when the ASR of the backdoor samples in the poisoned CIFAR-100 datasets are greater than $99\%$. Finally, we get the backdoor models with $99.94\%$, $99.90\%$, $73.37\%$ ASR on the poisoned ImageNet samples and with the performance degradation no more than $3\%$ on the CIFAR-10 task, as shown in Table~\ref{tab:Transformers in CIFAR-10}. Notably, the backdoor injection time is only $157$s, $4$s, and $195$s trained  on  1  Nvidia  GeForce  RTX  3090GPU. Furthermore, we also visualize some poisoned examples in (d)-(f) of Figure~\ref{fig:poison example}.

\section{Discussion}
\label{sec:discussion}
\noindent\textbf{Limitation.}
We proposed a data-free backdoor attack method against CV-oriented transformer models. We consider that there are some methods to provide the attention map for CNNs, such as Grad-CAM~\cite{selvaraju2017grad}, Grad-CAM++~\cite{Chattopadhyay2018GradCAMGG}. In order to test the generalization of DBIA, we train and test on two classic CNN models trained on ImageNet task: Resnet50\footnote{https://download.pytorch.org/models/resnet50-0676ba61.pth} and VGG19\footnote{https://download.pytorch.org/models/vgg19-dcbb9e9d.pth} officially launched by PyTorch, and the surrogate dataset is CIFAR-10. The experimental results show that, for Resnet50, when our attack success rate reaches $33.65\%$, the accuracy of the main task has decreased from $73.90\%$ to $61.55\%$. Accordingly, on VGG19, when the attack success rate reaches $33.20\%$, the accuracy of the main task decreases from 71.65\% to $53.65\%$. After analysis, we 
think the main reason of this result is that the interpretability of CNNs based on convolution computation is not as good as that of CV-oriented transformers based on attention mechanism.

\noindent\textbf{Adaptive Attack against Defense Methods.} 
None of the attacks can resist all backdoor defense methods, and we can further add the adaptive loss term into the backdoor injection loss to improve the resistance of our backdoor attack against specific defense methods. For example, both Februus~\cite{doan2020februus} and SentiNet~\cite{chou2018sentinet} aim to locate the trigger areas that the model focuses on by Grad-CAM~\cite{selvaraju2017grad}, and we can design the adaptive loss term to resist the detection of these defense methods as below: $loss_{ada}=loss(GCAM(\widetilde{x}, y_{t})) - mask)$. Where $mask$ represents the target position that we expect Grad-CAM focus, which does not coincide with the trigger position, we can obtain the backdoor to evade these defense methods by minimizing the loss. Similarly, we can design other adaptive loss terms against other specific defense methods accordingly to improve the resistance of the backdoor attacks.


\section{Conclusion}
In this paper, we studied the backdoor attack on CV-oriented transformers. We consider a more realistic backdoor attack scene: without access to the datasets related to the main tasks, most existing backdoor attacks are likely to fail under this condition. To address this, we propose the attention maximum trigger against the surrogate datasets based on the attention mechanism of transformer models to focus on the trigger in advance and then fine-tune a few neurons with Projected Gradient Descent to specify the focused trigger the target label. We show that our proposed attack can inject a backdoor into these transformers with a high attack success rate on three state-of-the-art CV-oriented transformers well-trained on two benchmark datasets.

{\small
\bibliographystyle{ieee_fullname}
\bibliography{egbib}
}

\end{document}